# UA-Code-Bench: A Competitive Programming Benchmark for Evaluating LLM Code Generation in Ukrainian

**Mykyta V. Syromiatnikov**[1)]
Postgraduate Student of the Department of Software Engineering
ORCID: orcid.org/0000-0002-0610-3639; nik.syromyatnikov@gmail.com
**Victoria M. Ruvinskaya**[1)]
PhD, Professor of the Department of Software Engineering
ORCID: orcid.org/0000-0002-7243-5535; ruvinska@op.edu.ua
[1)] Odesa Polytechnic National University, 1, Shevchenko Ave. Odesa, 65044, Ukraine

**ABSTRACT**

Evaluating the real capabilities of large language models in low-resource languages still represents a challenge, as many existing benchmarks focus on widespread tasks translated from English or evaluate only simple language understanding. This paper introduces UA-Code-Bench, a new open-source benchmark established for a thorough evaluation of language models' code generation and competitive programming problem-solving abilities in Ukrainian. The benchmark comprises 500 problems from the Eolymp platform, evenly distributed across five complexity levels from very easy to very hard. A diverse set of 13 leading proprietary and open-source models, generating Python solutions based on a one-shot prompt, was evaluated via the dedicated Eolymp environment against hidden tests, ensuring code correctness. The obtained results reveal that even top-performing models, such as OpenAI o3 and GPT-5, solve only half of the problems, highlighting the challenge of code generation in low-resource natural language. Furthermore, this research presents a comprehensive analysis of performance across various difficulty levels, as well as an assessment of solution uniqueness and computational efficiency, measured by both elapsed time and memory consumption of the generated solutions. In conclusion, this work demonstrates the value of competitive programming benchmarks in evaluating large language models, especially in underrepresented languages. It also paves the way for future research on multilingual code generation and reasoning-enhanced models.

**Introduction.** Large language models (LLMs), especially LLM-powered assistants, are now common tools in both general and expert work. Hundreds of millions use them for search, guidance, analysis, and everyday automation. Nowadays, even coding, which seemed to be an overly complex task before, is no longer a stopper for modern reasoning models that turn natural-language specifications into working code, build or refactor systems, write tests, and even review code. Tools like GitHub Copilot improve productivity (up to a 55% reduction in task completion time) and lower mental load for 73% of developers on repetitive coding tasks [1]. This indicates strong demand for AI assistance across various domains.

Despite these successes, most code generation benchmarks and validations have focused on problems described in English. Moreover, the studies have shown that LLM performance in low-resource languages degrades substantially on high-precision tasks [2]. However, the field of LLM benchmarking in the Ukrainian language mainly consists of standard datasets with trivial tasks like classification or question answering, as well as translated English evaluation sets. In addition, the rule of thumb is that the quality of the input directly influences the obtained result. For instance, when a student or engineer poses a coding challenge in the relatively low-resource Ukrainian language, even advanced code assistants may fail to respond, effectively forcing users to translate their problems into English, which is not always exact and slows work, widening the gap with English speakers. Support for local languages is therefore also about fairness and access.

To close the gap, this work aims to establish UA-Code-Bench – a first Ukrainian language-native code generation benchmark, enabling rigorous evaluation of LLM code generation correctness and efficiency on competitive programming tasks. This includes the following tasks:

- building, for the first time to our knowledge, a large-scale Ukrainian code-generation benchmark with programming tasks of different complexities sampled from the Eolymp platform;

---

- evaluating modern open and proprietary LLMs, with their correctness assessed using test suites from competitive-programming platforms such as Eolymp;
- analyzing overall and per-difficulty performance, as well as memory/latency trade-offs.

These steps shall provide insights into which models excel or struggle, and why, shedding light on the current limitations of AI in competitive coding and low-resource language understanding.

**Related works.** Automatic code generation has progressed through several phases of active development. Early 2000s systems were rule-based, fully transparent at a cost of poor generalization and evaluated by code-quality metrics such as cyclomatic complexity. Later, statistical machine learning redefined the task as probabilistic modeling, shifting evaluation toward similarity scores like BLEU. Pre-LLM neural methods yielded narrow tools like NL2Bash for short command synthesis [3] and code search from natural-language queries [4]. Invention of Transformer architecture enabled large-scale pretraining and tuning, leading to LLMs for code like Codex and AlphaCode, alongside open efforts such as StarCoder and compact open-weight families (CodeGemma, QwenCoder).

As Transformer-based models have grown more powerful, the benchmarking side of the research has evolved from somewhat naïve similarity-based metrics to metrics that evaluate the correctness of generated solutions. HumanEval and MBPP standardized unit-test metrics (e.g., pass@k measuring if any of those k solutions pass all unit tests) on short English prompts [5, 6]. APPS raised difficulty with tasks of varying complexity, from one-liners to substantial algorithmic challenges, and showed early large models solved only a small fraction of easy problems [7], motivating stronger approaches. Benchmarks like HumanEval, MBPP, and APPS primarily focused on isolated self-contained functions, while a day-to-day software development requires a lot more. To match real software work, SWE-bench moved from single functions to fixing issues in large codebases [8].

The big problem with static benchmarks is that models may have been inadvertently trained on solutions widely available online. Fresh works are mitigating the issue, continuously collecting new problems (LiveCodeBench) [9] and evaluating on a set of private test cases (CodeElo) [10].

While being methodologically superior, existing competitive programming benchmarks have focused exclusively on high-resource languages, such as English, leaving a critical gap in the evaluation of low-resource languages. At the same time, HumanEval-XL indicates that even powerful models like GPT-4 perform markedly worse in languages with less training data available [2], often called low-resource languages, such as Ukrainian.

In recent years, the Ukrainian NLP community has made significant contributions by developing evaluation resources. ZNO-Eval provides a benchmark for assessing the general capabilities of LLMs using tasks from standardized national school exams [11], and follow-up studies focused on reasoning models [12]. ZNO-Vision, in turn, extended text-only evaluation on school exams with multimodal understanding [13]. UAlign was introduced to evaluate the ethical alignment in a cultural context [14].

This growing ecosystem of benchmarks is crucial for the development of Ukrainian language models. However, a dedicated resource for evaluating code generation capabilities has long been absent. This underscores the novelty of UA-Code-Bench, designed to fill the gap.

**Benchmark methodology**. UA-Code-Bench is built from Eolymp, a widely used Ukrainian platform with Ukrainian-language statements and hidden test cases evaluated by an automated judging system. The benchmark contains 500 sampled problems, evenly split across five difficulty bands (from very easy to very hard) assigned by Eolymp based on algorithmic demands, coding intricacy, and typical acceptance rate. Lower bands cover basic math, text processing, and simple control flow. Medium items often combine several algorithms. Hard and very hard mirror contest problems and require not only intricate reasoning but also efficient implementations for large inputs under strict time and memory constraints.

Thirteen leading LLMs, proprietary and open-weight, general-purpose and code-specialized, were evaluated in a one-shot setting. Each prompt contained one Ukrainian example plus a short correct solution to guide the model on the expected input and output formats. For every task, the model produced a single solution without retries under a 30-minute timeout, using recommended sampling parameters with low temperature (0-0.2). The programming language was set to Python 3 for solution generation.

To judge solution correctness, an automated submission tool was implemented. In a dedicated space with a seat purchased for each model, this tool submitted code to the Eolymp judge, which executed it against private tests. A submission was accepted only if all tests passed, wrong answers earned partial credit, while exceptions, timeouts, or runtime failures counted as failures. Two primary metrics were used: pass@1 (accepted if 100% of hidden tests pass) and average score (0 to 100). Five secondary indicators complemented them: TOO (tasks uniquely solved by a model), T1T (accepted solutions with the smallest worst-case execution time), T1M (same as T1T, but for memory), GE (generation failure, including timeout/invalid or incomplete code), and EE (compile/runtime error). The benchmark, data parsing, preparation, code generation, and evaluation scripts are available at https://huggingface.co/datasets/NLPForUA/ua-code-bench.

**Results and Analysis**. During evaluation, it was discovered that out of 500 problems, 14 (1 hard, 13 very hard) failed due to grader issues and were removed. Table 1 presents evaluation results.

*Table 1*. **Overall code generation result per model (486 tasks total): TOO – number of tasks uniquely solved by a model; T1T – number of accepted solutions with the smallest worst-case execution time; T1M – same as T1T, but for memory consumption; GE – generation error; EE – execution error.**

| Model | Params, billions | Accepted solutions | Average score (%) | TOO | T1T | T1M | GE | EE |
|---|---|---|---|---|---|---|---|---|
| MamayLM 9b | 9 | 16 | 11,97 | 0 | 1 | 3 | 1 | 36 |
| GPT-OSS-20b low | 20 | 158 | 50,81 | 0 | 7 | 10 | 2 | 20 |
| GPT-OSS-20b medium | 20 | 208 | 61,50 | 1 | 19 | 10 | 3 | 13 |
| Gemma-3-27b-it | 27 | 53 | 25,70 | 0 | 7 | 6 | 3 | 11 |
| Qwen2.5-Coder-32b-Instruct | 32 | 60 | 26,63 | 0 | 0 | 5 | **0** | **1** |
| GPT-OSS-120b low | 120 | 188 | 57,66 | 1 | 7 | 3 | 11 | 2 |
| GPT-OSS-120b medium | 120 | 219 | 65,99 | 2 | 5 | 6 | 3 | 3 |
| DeepSeek-R1-0528 | 671 | 198 | 61,33 | 0 | 14 | 6 | 15 | 3 |
| Grok 3 | N/A | 96 | 40,48 | 1 | 1 | 18 | **0** | 2 |
| Grok 4 | N/A | 172 | 45,91 | 2 | 5 | 21 | 190 | 2 |
| Claude Opus 4 | N/A | 158 | 57,51 | 0 | 11 | 17 | 1 | 2 |
| Gemini 2.5 pro | N/A | 207 | 61,96 | 3 | 25 | 37 | 42 | **1** |
| OpenAI o4-mini medium | N/A | 238 | **68,05** | 5 | 28 | 19 | 2 | 10 |
| OpenAI o3 medium | N/A | **246** | 67,60 | 3 | **44** | 26 | 16 | 2 |
| OpenAI GPT-5 medium | N/A | 244 | 66,50 | **12** | 15 | **47** | 57 | 4 |

The top group, with OpenAI o3 (246 accepted), OpenAI GPT-5 (244), and OpenAI o4-mini (238), all with medium reasoning effort, substantially outperforms the others. However, even the best system leaves 242 tasks unsolved. About half of all problems (mainly hard/very hard) remain unsolved by any model. Figure 1 shows per-difficulty stats, and Figure 2 visualizes the average score per model, accounting for partially correct solutions, which pass some, but not all, of the hidden test cases. As Figures 1-2 show, the performance drops sharply as task complexity moves from easy to hard. This indicates that while many models can perform direct translation of simple Ukrainian instructions into code, they lack the deep algorithmic reasoning. In general, the obtained results align with model size and prior English-benchmark results [15].

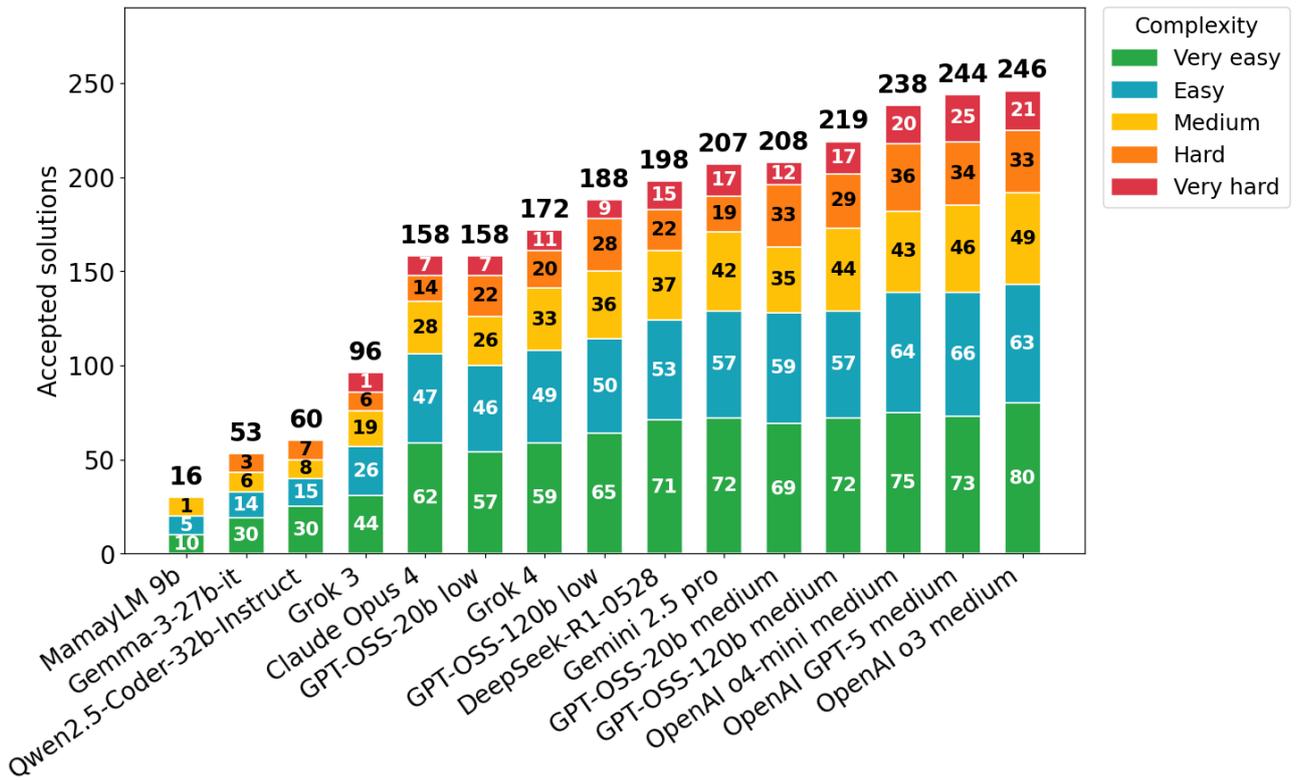

*Fig. 1.* **Number of accepted solutions per model, broken down by problem difficulty. Each bar sums to the total solved by that model (max 486). Bars are distorted on y-axis for readability**

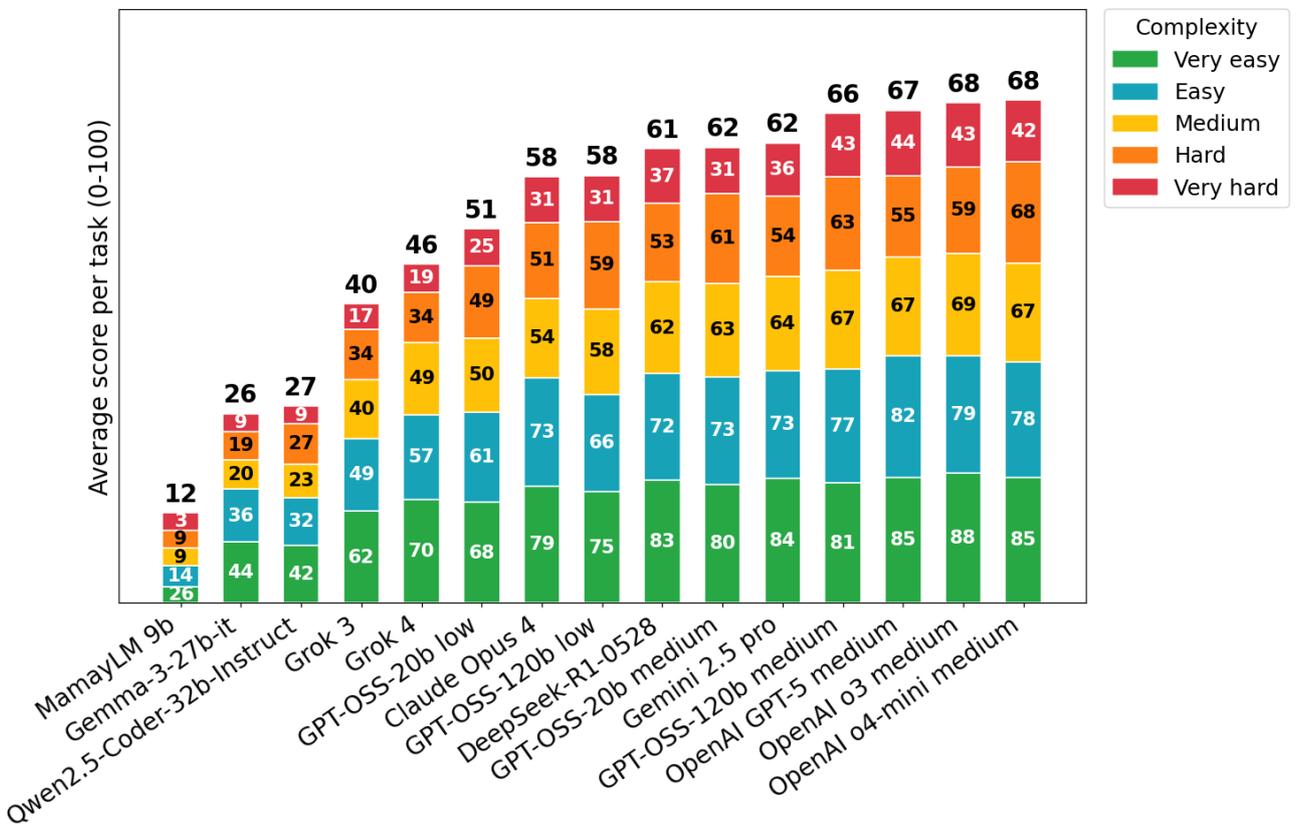

*Fig. 2.* **Average scores for the task per model, broken down by problem difficulty. Each bar represents the average score by model (max 100). Bars are distorted on y-axis for readability**

An important indicator of a model's advanced reasoning is its ability to solve problems that no other model can. Figure 3 presents a chart of how many tasks were uniquely solved by each model.

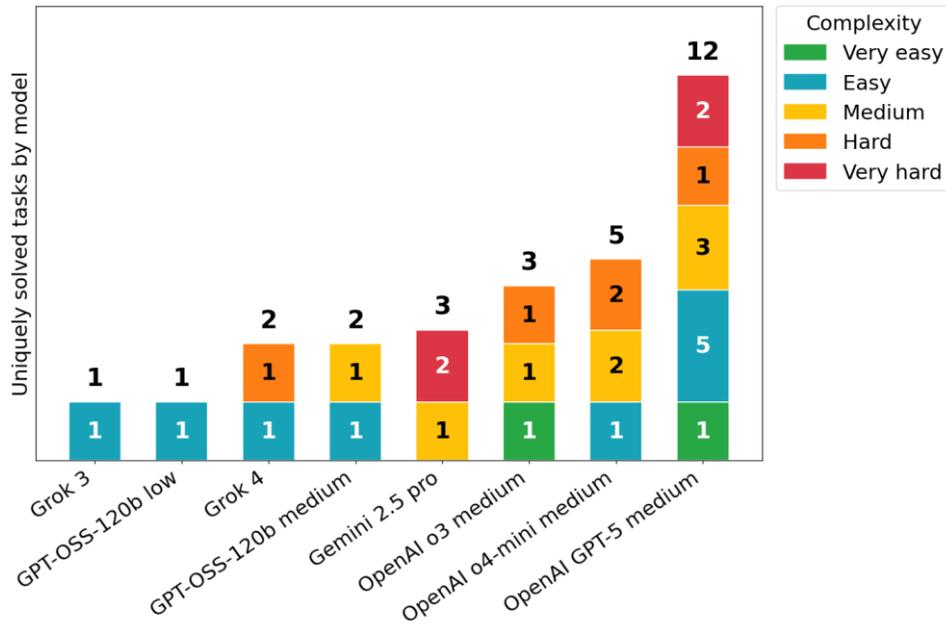

*Fig. 3.* **Distribution of uniquely solved tasks by model and complexity**

The standout performer in this category is GPT-5 with 12 unique tasks. The small number of unique solutions indicates heavy overlap, as LLMs tend to succeed on the same set of simple problems.

In competitive programming, a correct solution is often insufficient. Fig. 4 and Fig. 5 present an analysis of accepted submissions with smallest worst-case execution time and memory.

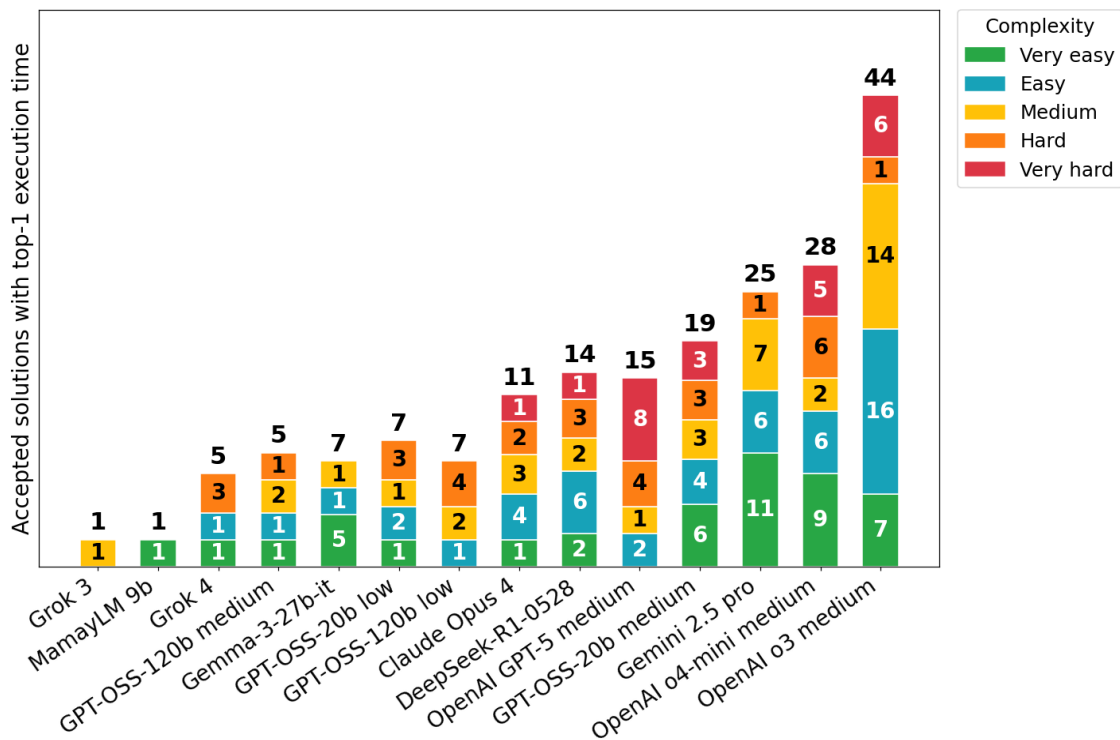

*Fig. 4.* **Number of accepted solutions with the smallest worst-case execution time (bars are distorted on y-axis for readability)**

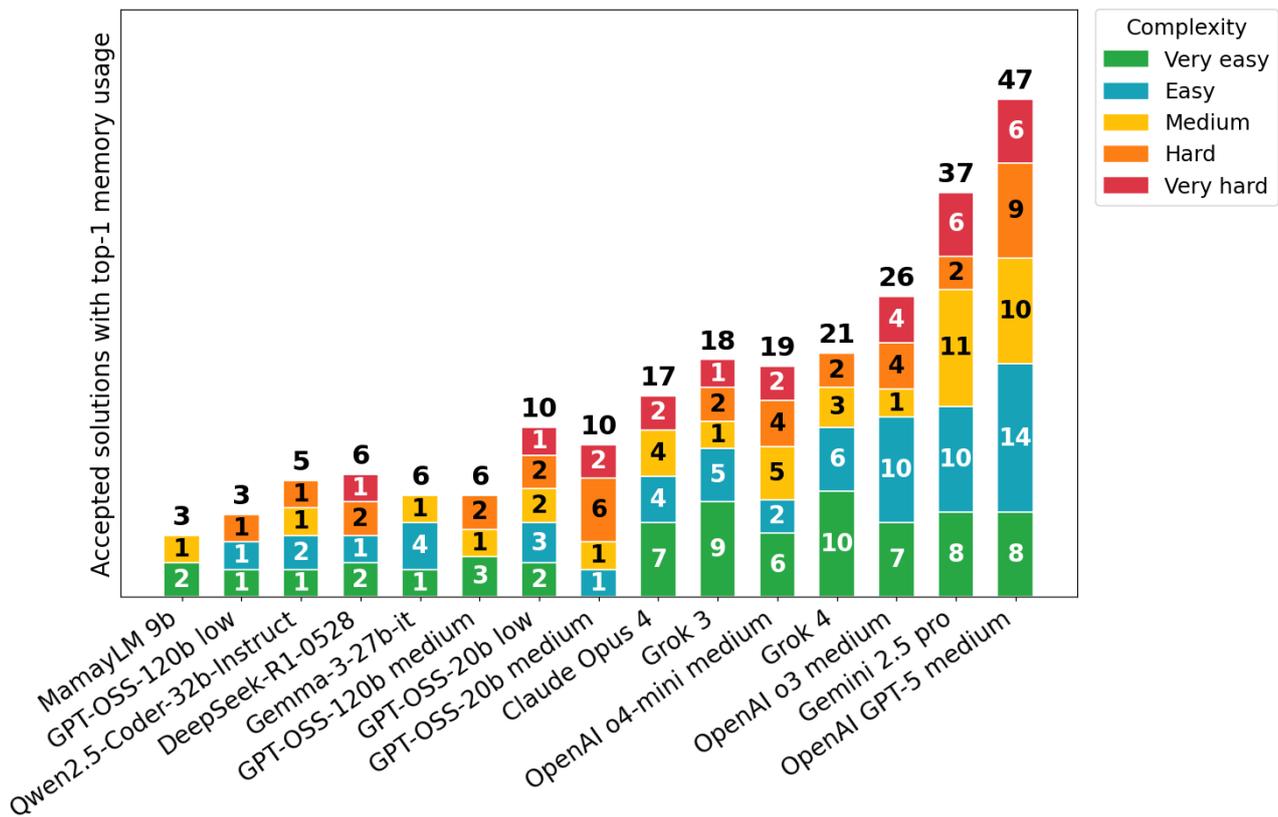

*Fig. 5*. **Number of accepted solutions with the smallest worst-case memory consumption (bars are distorted on y-axis for readability)**

OpenAI o3 is the speed champion on 44 tasks, while GPT-5 produces the most memory-efficient code in 47 instances. This trade-off between speed and memory highlights that there is no single best model since models exhibit distinct strengths that may be more or less desirable depending on the specific application constraints. Interactive evaluation results can be accessed at https://uallm.org/code-evaluation.

**Conclusion.** This research presented UA-Code-Bench, the first competitive programming benchmark designed to evaluate the code generation and reasoning capabilities of large language models in the Ukrainian language. The evaluation of 13 modern LLMs on 500 problems from the Eolymp platform conducted by this research leads to several key takeaways:

- large LLMs can solve complex Ukrainian tasks, but the performance of all models degraded sharply with increasing task complexity;
- model intelligence and generalization ability in low-resource languages still have not reached their peak, even for the largest solutions, so more work on multilingual training is needed to close the gap;
- competitive programming tasks meaningfully stretch evaluated models, exposing weaknesses that simpler benchmarks may hide.

The limitations of this research include the narrowness of the benchmark, with only one programming language being tested (Python), and a single source of problem tasks (the E-olymp platform), without an in-depth analysis of data contamination. Future work should focus on diversity by extending benchmarks through the addition of more problems, including multi-modal tasks that incorporate joint text and visual information, and by evaluating a wider range of models. Additionally, thorough task and submission analysis should be conducted, including fine-grained error categories.

Finally, this work emphasizes that code generation presents a dual challenge, requiring both natural language understanding and the generation of correct and efficient code. Hopefully, UA-Code-Bench, which stresses both aspects, will serve as a valuable resource for evaluating the next generation of AI coding assistants, ensuring they become not only smarter in general but also more inclusive and effective across languages and cultures.

## ACKNOWLEDGEMENT

We gratefully acknowledge the Eolymp team for maintaining a robust platform, online judging infrastructure, and high-quality Ukrainian problem statements with hidden tests, without which this study would not have been possible. We also thank the organizations, companies, and open-source communities that release open models and tools, whose openness substantially advances research reproducibility and accessibility.

# UA-Code-Bench: україномовний бенчмарк спортивного програмування для оцінювання генерації коду великими мовними моделями

Сиром'ятніков Микита Валерійович[1)]
Аспірант каф. Інженерії програмного забезпечення
ORCID: orcid.org/0000-0002-0610-3639; nik.syromyatnikov@gmail.com

Рувінська Вікторія Михайлівна[1)]
Доктор філософії, професор каф. Інженерії програмного забезпечення
ORCID: orcid.org/0000-0002-7243-5535; ruvinska@op.edu.ua

[1)] Національний університет «Одеська політехніка», пр. Шевченка, 1. Одеса, 65044, Україна

## АНОТАЦІЯ

Оцінювання реальних можливостей великих мовних моделей у низькоресурсних мовах все ще залишається складним завданням, оскільки значна частина наявних тестових наборів даних зосереджуються на поширених задачах, перекладених з англійської, або перевіряють лише базове розуміння мови. У цій роботі представлено UA-Code-Bench – новий загальнодоступний бенчмарк для всебічного оцінювання здатності великих мовних моделей генерувати програмний код і розв'язувати україномовні задачі зі спортивного програмування. Набір охоплює 500 задач платформи Eolymp, рівномірно розподілених за п'ятьма рівнями складності – від дуже простих до дуже складних. Різноманітний набір із 13 провідних пропрієтарних та загальнодоступних великих мовних моделей, що генерували код рішення на Python за інструкцією із одним прикладом (one-shot), було оцінено у виділеному середовищі Eolymp на прихованих тестах, що перевіряють правильність рішення. Отримані результати демонструють, що навіть найкращі моделі, зокрема OpenAI o3 та GPT-5, розв'язують лише половину задач. Це підкреслює складність генерації коду для умов, описаних низькоресурсною мовою. Додатково представлено детальний аналіз продуктивності за рівнями складності, а також оцінювання унікальності розв'язків і ефективності згенерованих рішень, що оцінювалася швидкістю виконання та споживанням пам'яті згенерованих програм. Підсумовуючи, робота демонструє цінність діагностичних наборів даних зі спортивного програмування для оцінювання великих мовних моделей, особливо для перевірки здібностей у низькоресурсних мовах, і окреслює шлях до подальших досліджень багатомовної генерації коду та моделей з підтримкою міркування.

**Ключові слова:** великі мовні моделі, генерація коду, бенчмарк, спортивне програмування, українська мова